\documentclass[10pt,twocolumn,letterpaper]{article}

\usepackage{cvpr}              

\usepackage[dvipsnames]{xcolor}

\usepackage{times}
\usepackage{epsfig}
\usepackage{graphicx}
\usepackage{amsmath}
\usepackage{amssymb}
\usepackage{booktabs}
\usepackage{diagbox}

\usepackage{url}            
\usepackage{amsfonts}       
\usepackage{nicefrac}       
\usepackage{microtype}      
\usepackage{xcolor}         

\usepackage{subcaption}
\usepackage{tikz}
\usepackage{comment}
\usepackage{xcolor}
\usepackage{siunitx,caption}
\usepackage{subcaption}
\usepackage{pifont}
\usepackage{multirow}
\usepackage{makecell}
\usepackage{bold-extra}
\usepackage{listings}
\usepackage{algorithm}
\usepackage{bm}
\usepackage{newtxtext}
\usepackage[accsupp]{axessibility} 
\usepackage[
singlelinecheck=false 
]{caption}

\usepackage{threeparttable}

\newcommand{\ModelName}{LoCoNet}
\newcommand{\cmark}{\ding{51}}%
\newcommand{\xmark}{\ding{55}}%
\newcommand{\xhdr}[1]{\vspace{4pt} \noindent {\textbf{#1}}}

\definecolor{cvprblue}{rgb}{0.21,0.49,0.74}
\definecolor{arylideyellow}{rgb}{0.91, 0.84, 0.42}
\definecolor{asparagus}{rgb}{0.53, 0.66, 0.42}
\definecolor{aureolin}{rgb}{0.99, 0.93, 0.0}
\definecolor{buff}{rgb}{0.94, 0.86, 0.51}
\definecolor{ao}{rgb}{0.0, 0.5, 0.0}
\usepackage[pagebackref,breaklinks,colorlinks,citecolor=cvprblue]{hyperref}

\def\paperID{4519} 
\def\confName{CVPR}
\def\confYear{2024}

\title{LoCoNet: \textit{Lo}ng-Short \textit{Co}ntext \textit{Net}work for Active Speaker Detection}

\author{Xizi Wang$^1$ \quad Feng Cheng$^2$ \quad Gedas Bertasius$^2$ \quad David Crandall$^1$\\
 $^1$Indiana University \quad $^2$UNC Chapel Hill \\
{\tt\small xiziwang@iu.edu \quad \{fengchan,gedas\}@cs.unc.edu}
}
\begin{document}
\maketitle
\begin{abstract}
   Active Speaker Detection (ASD) aims to identify who is speaking in each frame of a video. Solving ASD involves using audio and visual information in two complementary contexts: long-term intra-speaker context models the temporal dependencies of the same speaker, while short-term inter-speaker context models the interactions of speakers in the same scene. Motivated by these observations, we propose \ModelName, a simple but effective Long-Short Context Network that leverages Long-term Intra-speaker Modeling (LIM) and Short-term Inter-speaker Modeling (SIM) in an interleaved manner. LIM employs self-attention for long-range temporal dependencies modeling and cross-attention for audio-visual interactions modeling. SIM incorporates convolutional blocks that capture local patterns for short-term inter-speaker context. 
Experiments show that \ModelName~achieves state-of-the-art performance on multiple datasets, with 95.2\% (+0.3\%) mAP on AVA-ActiveSpeaker, 97.2\% (+2.7\%) mAP on Talkies, and 68.4\% (+7.7\%) mAP on Ego4D. Moreover, in challenging cases where multiple speakers are present, \ModelName~outperforms previous state-of-the-art methods by 3.0\% mAP on AVA-ActiveSpeaker. The code is available at \url{https://github.com/SJTUwxz/LoCoNet_ASD}.
\end{abstract}    
\section{Introduction}
\label{sec:intro}

Real-world interactive computer vision systems need to recognize not only the physical properties of a scene, such as objects and people, but also the social properties, including how people interact with each other.
One  fundamental task is identifying, at any moment, who is speaking in a complex scene with multiple interacting individuals. This Active Speaker Detection (ASD) problem~\cite{alcazar2020active,kopuklu2021design,tao2021someone,grauman2022Ego4D,leon2021maas,alcazar2022end,roth2020ava,chakravarty2016cross,kim2021look,jung2023talknce,gurvich2023real,kyoung2023modeling,huang2020improved,everingham2009taking,qian2021audio,saenko2005visual,slaney2000facesync,wuerkaixi2022rethinking} is crucial for many real-world applications like human-robot interaction~\cite{thomaz2016computational,kang2021video,skantze2021turn}, speech diarization~\cite{garcia2017speaker,wang2018speaker,xu2021ava,gebru2017audio,ding2020self,chung2020spot}, video re-targeting~\cite{bansal2018recycle,li2018depth,yang2020transmomo}, multimodal learning~\cite{beyan2020realvad,shahid2021s,gao2020listen,kazakos2019epic,hu2021class,rao2022decompose,arandjelovic2018objects,son2017lip,jati2019neural,owens2018audio,michelsanti2021overview,duan2022flad,kazakos2019epic,ngiam2011multimodal,owens2018audio,qian2021multi,rao2022decompose}, etc.

\setlength{\belowcaptionskip}{-5pt}
\begin{figure}[t]
\centering
\includegraphics[width=\linewidth]{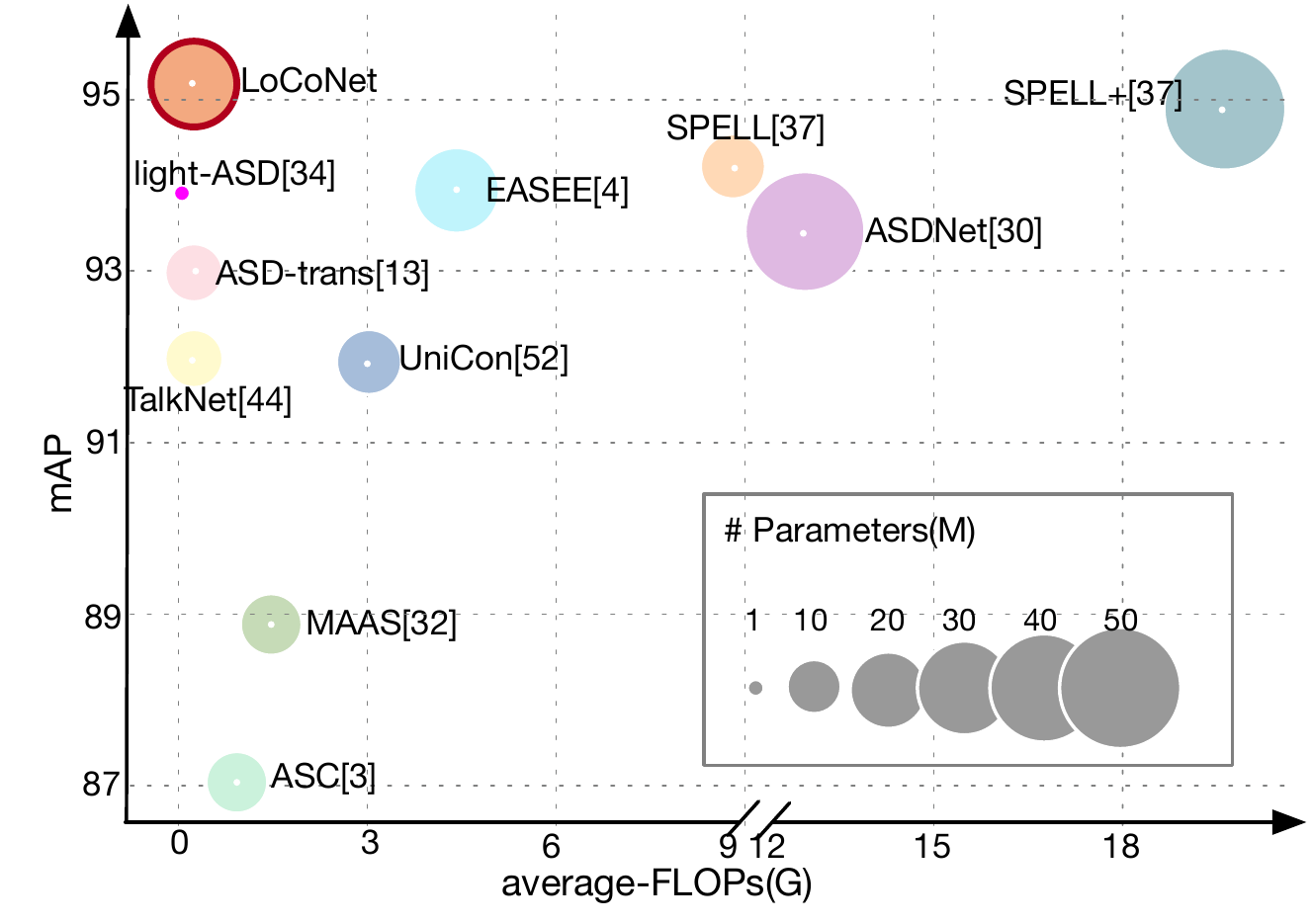}
\caption{\textbf{Comparison of ASD methods in terms of mean average precision (mAP) on the AVA-ActiveSpeaker dataset, average-FLOPs, and number of parameters.} Note that average-FLOPs is the computation required to predict the speaking activity of one face crop.}

\label{fig:comparison}
\end{figure}

\begin{figure*}[ht]
\centering
\includegraphics[width=\textwidth]{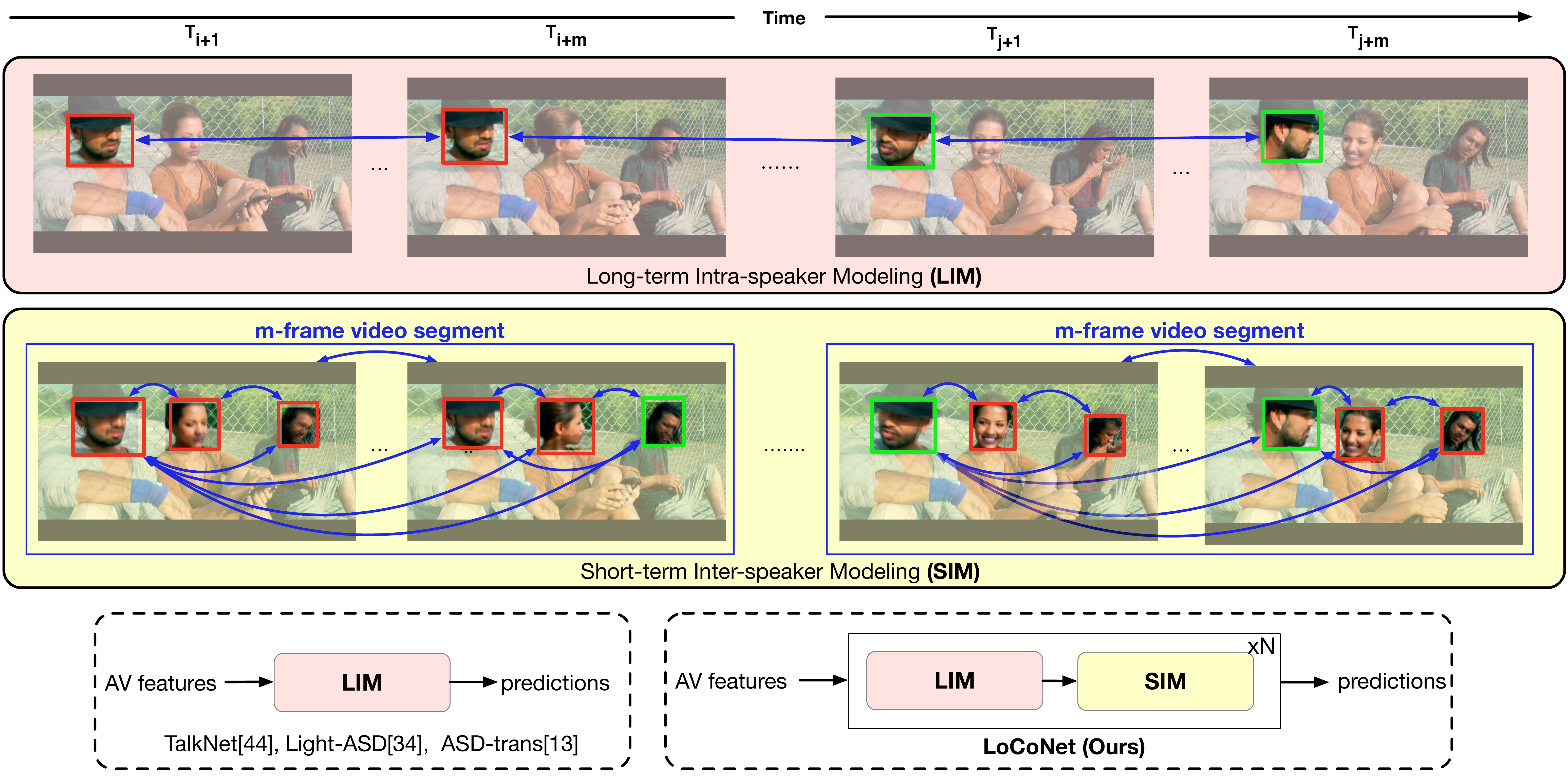}
\caption{\textbf{Long-term Intra-speaker Modeling (\textcolor{pink}{LIM}), Short-term Inter-speaker Modeling (\textcolor{buff}{SIM}), and comparison of \ModelName~with existing long-term parallel-inference ASD methods}. \textcolor{red}{Red} boxes show  inactive speakers and \textcolor{green}{green} boxes show active speakers. LIM uses the features of a single speaker across all frames to capture long-term relationships. SIM models the relationships of speakers within a short $m$-frame segment to capture the conversation pattern. The speaker context modeling of the existing long-term parallel-inference ASD methods~\cite{tao2021someone,liao2023light,datta2022asd} only focuses on LIM, while \ModelName~models LIM and SIM to learn both contexts.}

\label{fig:intro}
\end{figure*}

How can we tell whether someone is speaking?
Visual cues such as movements of the mouth and eyes, when correlated with audio signals, often offer direct and primary evidence. 
 The inter-modality synchronization over a longer audio-visual segment also provides complementary   information~\cite{tao2021someone,alcazar2020active,min2022learning,kopuklu2021design}.  
The first row of Fig~\ref{fig:intro} shows how Long-term Intra-speaker Modeling  helps discern this primary indicator by observing one person for a long time span.

However, in a complex video, a person's face is often occluded, turned away,  off-frame, or very small, posing challenges for directly inferring speaking activity from visual cues.
Fortunately, valuable evidence about the target speaker can be gleaned from the behaviors of others in the scene~\cite{skantze2021turn}.
The second row of Fig~\ref{fig:intro} shows an example: from the left $m$-frame video segment, even with a partially visible face of the man on the right, it is easy to see  that he is speaking at $T_{i+m}$ because the woman in the middle turns her head ($T_{i+1} \to T_{i+m}$) and neither of the other two people open their mouths at $T_{i+m}$. 
Notably, the woman's gaze towards the man on the right at $T_{i+m}$ does not provide substantial information on whether the man is speaking at a distant time $T_{j+1}$. 
Therefore, we argue that Inter-speaker Modeling is sufficient in short-term temporal windows, since  activities of speakers within a short time range are more correlated than speakers separated farther away in time. Cognitive research~\cite{richardson2008synchrony,levinson2015timing} also suggests that speaker-listener coupling is more coordinated in nearby frames.

To solve this per-frame video classification task, existing ASD methods can be put into two categories: 1) \textbf{parallel-inference} methods~\cite{tao2021someone,liao2023light,datta2022asd,zhang2021unicon,kyoung2023modeling} take all frames as input and predict the results for all frames in one pass. 2) \textbf{sequential-inference} methods~\cite{kopuklu2021design,alcazar2020active,alcazar2022end,min2022learning,leon2021maas} take a short clip as input and only give prediction result of the center frame. Thus, a sliding window strategy is often adopted to get results for all the frames. As depicted in Fig.~\ref{fig:comparison}, the average-FLOPs of parallel-inference methods~\cite{tao2021someone,datta2022asd,liao2023light} to predict the speaking activity of one face crop are often much lower than sequential-inference methods~\cite{kopuklu2021design,alcazar2020active,leon2021maas,alcazar2022end,min2022learning}. However, most parallel-inference methods~\cite{tao2021someone, datta2022asd,liao2023light, xiong2022look} that take long video clips as input do not consider multi-speaker context, which could cause performance degradation as speakers' interaction is crucial for ASD task.

With the above issues in mind, we propose \ModelName, an end-to-end Long-Short Context Network. Long-term Intra-speaker Modeling (LIM) employs a self-attention mechanism~\cite{vaswani2017attention} for long-range dependencies modeling and a cross-attention mechanism for audio-visual interactions. Short-term Inter-speaker Modeling (SIM) incorporates convolutional blocks to capture local conversational patterns. 
Moreover, while most ASD methods use vision backbones for audio encoding~\cite{alcazar2020active,tao2021someone,leon2021maas,alcazar2022end} due to high temporal downsampling in most audio backbones~\cite{kong2020panns,hershey2017cnn,ravanelli2018speaker,chen2020vggsound,gong2021ast}, we propose VGGFrame to leverage pretrained AudioSet~\cite{gemmeke2017audio} weights to extract per-frame audio features. We also use a parallel inference strategy for more efficient video processing.

Our extensive experiments validate
the effectiveness of our approach.
On the AVA-ActiveSpeaker dataset~\cite{roth2020ava}, LoCoNet achieves 95.2\% mAP, outperforming previous state-of-the-art method SPELL+~\cite{min2022learning} by 0.3\% with $38\times$ less  computational cost. Furthermore, \ModelName~achieves  97.2\% (+2.7\%) mAP on  Talkies~\cite{leon2021maas} and 68.4\% (+7.7\%) mAP on  Ego4D's Audio-Visual benchmark~\cite{grauman2022Ego4D}. \ModelName~works especially well in challenging scenarios such as  with multiple speakers or small speaker faces.

\section{Related Work}
\label{sec:relatedwork}

Most recent techniques for ASD can be characterized in terms of three salient dimensions: frame-level processing strategy that determines the inference speed of the method, the extracted context information to enhance the feature representations for prediction, and the training mechanisms.

\xhdr{Frame-level processing strategy.}
Given a long video, existing ASD methods employ two main strategies for generating per-frame predictions. 1) \textit{Parallel-inference}~\cite{tao2021someone,liao2023light,datta2022asd,kyoung2023modeling} takes all frames of the video as input and predict the per-frame results in one pass. Such methods are often  fast, as depicted in Fig.~\ref{fig:comparison}. However, they typically do not consider interactions among multiple speakers.  2) \textit{Sequential-inference}~\cite{kopuklu2021design,alcazar2020active,leon2021maas,alcazar2022end,min2022learning} predicts one frame by sampling a short clip centered around that frame. They need a sliding window strategy to produce predictions for all frames, resulting in slower inference speed. Our proposed \ModelName~adopts a parallel inference strategy, combining fast inference speed with effective consideration of speakers' interactions.

\xhdr{Context Modeling.}
ASD  benefits from both intra-speaker and inter-speaker contexts. TalkNet~\cite{tao2021someone} models long-term temporal intra-speaker context to distinguish speaking and non-speaking frames. ASC~\cite{alcazar2020active} employs self-attention for long-term inter-/intra-speaker modeling, and an LSTM for long-term temporal refinement. ASDNet~\cite{kopuklu2021design} aggregates short-term features of target speaker and background speakers at nearby frames, and leverages Bidirectional GRU~\cite{cho2014properties} for long-term temporal modeling. Light-ASD~\cite{liao2023light} also uses Bidirectional GRU for temporal modeling.
MAAS~\cite{leon2021maas}, EASEE~\cite{alcazar2022end}, and SPELL~\cite{min2022learning} use Graph Convolutional Networks~\cite{wang2019dynamic,kipf2016semi} to model relationships between the visual nodes and audio nodes of context speakers. TS-TalkNet~\cite{jiang2023target} explores the use of reference speech to assist ASD.
 
In our proposed \ModelName, we introduce Long-Short Context Modeling (LSCM), composed of Long-term Intra-speaker Modeling (LIM) and Short-term Inter-speaker Modeling (SIM) in an interleaved manner. LIM captures long-term temporal dependencies with self-attention, and audio-visual interactions with cross-attention. SIM incorporates inter-speaker convolutional blocks to learn speakers' interactions.

\xhdr{Training mechanisms.}
Training on long videos can be memory-intensive, prompting some prior work~\cite{alcazar2020active,kopuklu2021design,leon2021maas,min2022learning} to adopt a
multi-stage training mechanism. In this process, a short-term feature extractor is initially trained, and subsequently, a long-term context modeling network is trained on the features extracted by the pre-trained feature extractor. 
TalkNet~\cite{tao2021someone}, EASEE~\cite{alcazar2022end}, UniCon~\cite{zhang2021unicon}, light-ASD~\cite{liao2023light}, ASD-transformer~\cite{datta2022asd}, and ADENet~\cite{xiong2022look} use end-to-end training, fully leveraging the learning capabilities of the model. Similarly, our proposed \ModelName~is also trained end-to-end, enabling joint optimization of audio-visual feature representation learning and context modeling.

\section{LoCoNet}
\label{sec:loconet}

\setlength{\belowcaptionskip}{-10pt}
\begin{figure}[t]
\centering
\includegraphics[width=\linewidth]{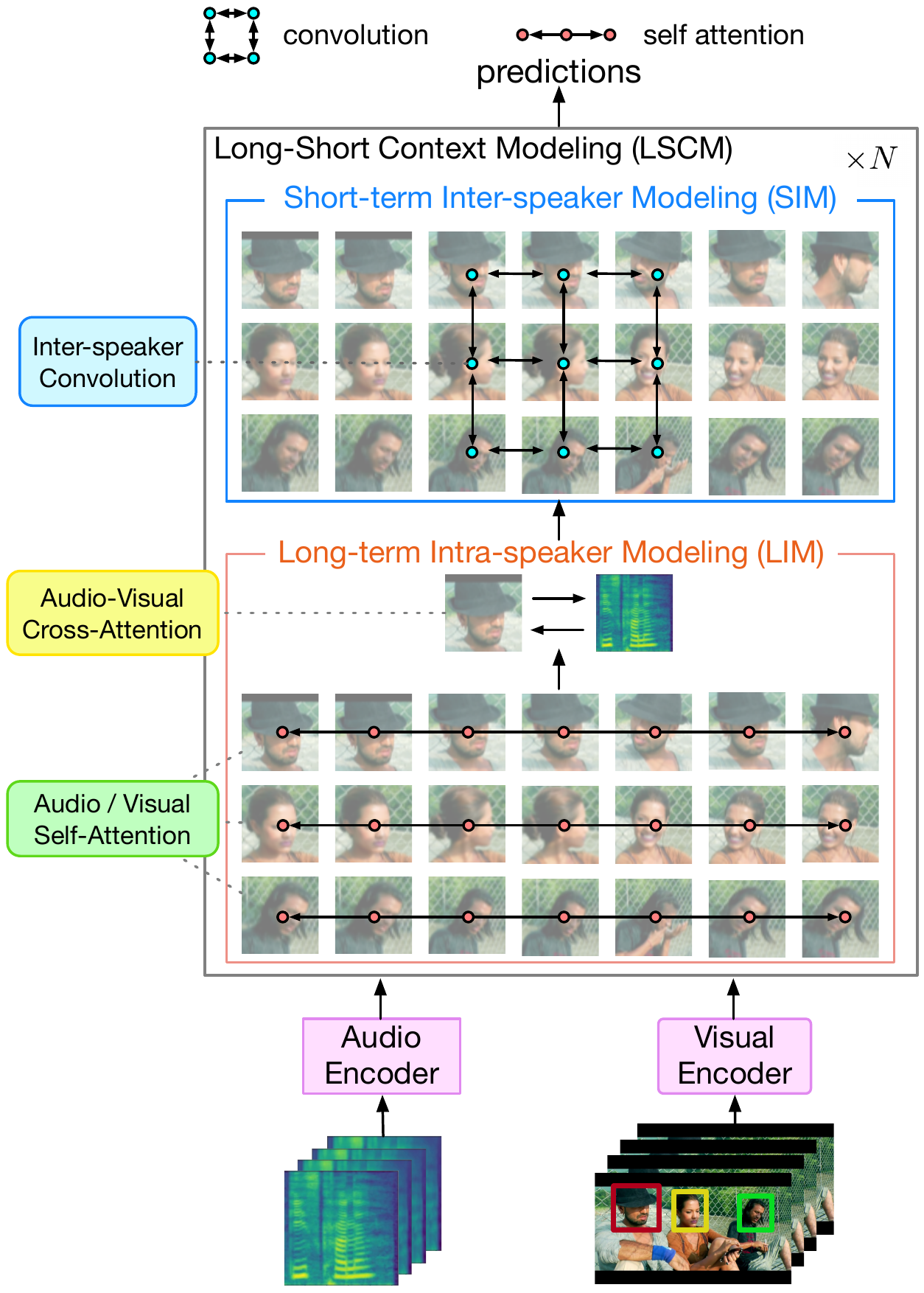}
\caption{\textbf{An overview of LoCoNet.} Given a sequence of face tracks and audio of a target speaker, we sample $S-1$ speakers from all other people appearing in the  scene and stack their face crops as visual input. Our method consists of 3 components: an audio encoder, a visual encoder, and a Long-Short Context Modeling module (LSCM) with $N$ blocks, where each block includes an attention-based Long-term Intra-speaker Model (LIM) and a convolution-based Short-term Inter-speaker Model (SIM) for speaker interaction. LIM involves Audio-Visual Self-Attention for long-term intra-speaker dependencies and Audio-Visual Cross-Attention for audio-visual interaction. The final output is used to classify speaking activity of the target person across all frames.}
\label{fig:face-audio}
\end{figure}

Given the stacked visual face track $V \in \mathbb{R}^{S\times T\times H \times W \times 1}$ and the audio Mel-spectrograms $A \in \mathbb{R}^{4T\times M}$, \ModelName~aims to predict the speaking activity $\hat{R} \in \mathbb{R}^{T}$ of the target person in each frame. $S$ is the number of speakers including the target speaker and $S-1$ context speakers in the same scene. $T$ is the temporal length of the face track. $H$ and $W$ are the height and width of each visual face crop. $M$ is the number of frequency bins of the audio Mel-spectrograms.

As shown in Fig~\ref{fig:face-audio}, \ModelName~consists of a visual encoder, an audio encoder, and a Long-Short Context Modeling (LSCM) module with $N$ LSCM blocks. We explain each module in more detail below.

\subsection{Encoders}
\xhdr{Visual encoder.} Given the face crop track $V_i \in \mathbb{R}^{T\times H\times W\times 1}$ of speaker $v_i$, the visual encoder yields a time sequence of visual embeddings $f_{v_i} \in \mathbb{R}^{T \times C}, i=1,...,S$. The stacked visual embeddings of the target speaker and all the sampled context speakers $f_{v} \in \mathbb{R}^{S \times T \times C}$ represent temporal context of each speaker independently.

\begin{figure}
\centering
\includegraphics[width=\linewidth]{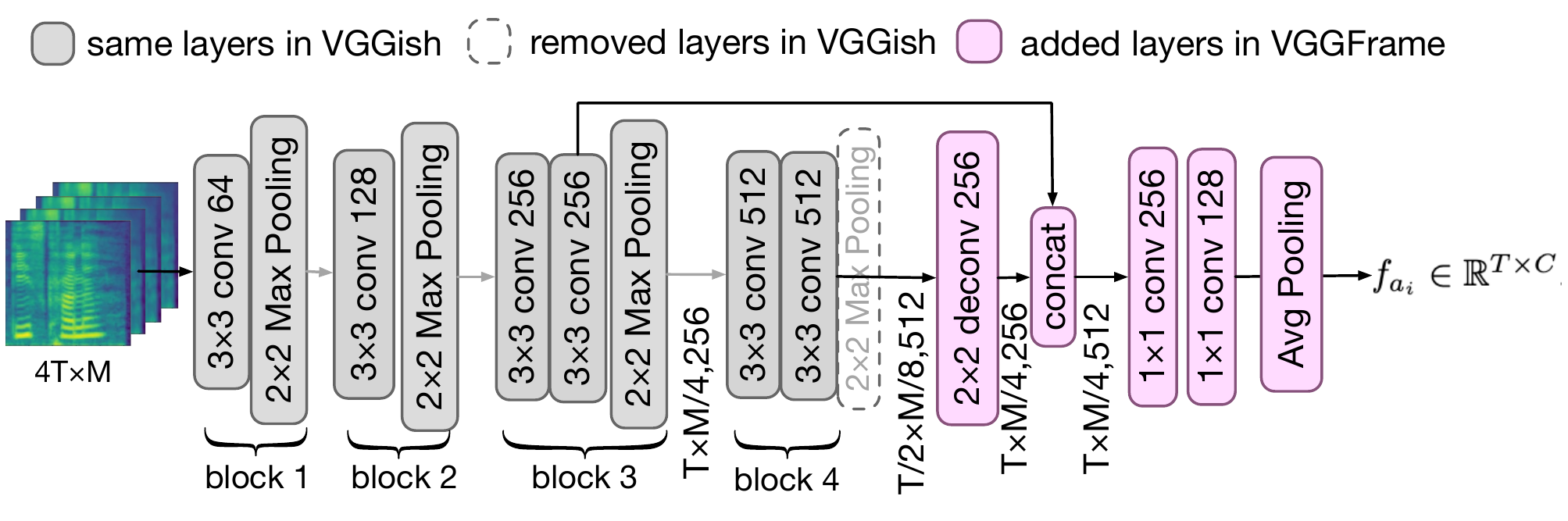}
\caption{An illustration of our proposed audio encoder VGGFrame. We apply a deconvolutional layer to upsample the output feature of block-4. The output features of block-3 (before max pooling) and deconvolutional layer are concatenated and transformed to per-frame output features of shape $T \times C$.}
\label{fig:audio-encoder}
\end{figure}

\xhdr{Audio Encoder.}
The audio encoder takes audio Mel-spectrograms $A \in \mathbb{R}^{4T\times M}$ as input. We need frame-level audio features for per-frame classification, but most pretrained audio encoders~\cite{hershey2017cnn,ravanelli2018speaker,chen2020vggsound,gong2021ast} are for audio classification and thus have a high degree of  temporal downsampling.
%
To solve this problem, we propose VGGFrame as the audio encoder, which can fully utilize the pretrained VGGish~\cite{hershey2017cnn}.
The architecture of VGGFrame is illustrated in Fig~\ref{fig:audio-encoder}. We remove the temporal downsampling layer after block-4 and add a deconvolutional layer to upsample the temporal dimension. We concatenate the intermediate features with the upsampled features to extract a hierarchy of representations.
VGGFrame outputs the audio embeddings $f_{a_i} \in \mathbb{R}^{T \times C}$. To align with $S$ speakers, we repeat  $f_{a_i}$ $S$ times to produce the audio embedding $f_a \in \mathbb{R}^{S\times T\times C}$.

\subsection{Long-Short Context Modeling}
\label{sec:lscm}

The visual and audio embeddings, derived independently by the visual and audio encoders, lack consideration of intra/inter-speaker context. Our Long-Short Context Modeling (LSCM) is designed to enhance these embeddings by learning  long-term intra-speaker and short-term inter-speaker context in an interleaved manner. As shown in Fig.~\ref{fig:face-audio}, LSCM consists of $N$ blocks, each incorporating a Long-term Intra-speaker Modeling (LIM) module and a Short-term Inter-speaker Modeling (SIM) module consecutively. 
LIM limits the model to look at the same speaker across all frames, encouraging it to learn speaker-independent  patterns from audio and visual interactions. In contrast, SIM constrains the model to examine all speakers in nearby frames and capture local interactions. Such inductive bias is instrumental in enhancing the model's capacity to glean valuable insights from these contextual dimensions.

LSCM inputs  audio embeddings $f_a$ and visual embeddings $f_v$, and produces context-aware embeddings $u_a^N, u_v^N \in \mathbb{R}^{S\times T\times C}$. These embeddings are concatenated to yield the final embeddings $u^N=concat(u_a^N, u_v^N)$. Via a linear layer, $u^N$ is used to predict the speaking activities $\hat{R} \in \mathbb{R}^{T}$ of the target person. The computation process at each LSCM block $l$ is detailed below.

\subsubsection{Long-term Intra-speaker Modeling (LIM)}
LIM consists of two submodules: (i) \textbf{Audio/Visual Self-Attention} models an individual person's behavior over a longer time period, and (ii) \textbf{Audio-Visual Cross-Attention} learns the interaction between audio and visual embeddings.

\xhdr{Audio/Visual Self-Attention.}
Since the model needs large capacity and to learn long-term dependencies, we employ the attention mechanism~\cite{vaswani2017attention} with a Transformer layer applied on the temporal dimension to achieve long-term modeling,
\begin{align} \label{eq1}
\tilde{u}_{v}^{l} &= \text{LN}(\text{MHA}(u_{v}^{l-1},u_{v}^{l-1},u_{v}^{l-1})+u_{v}^{l-1}),\\
\tilde{u}_{v}^{l} &= \text{LN}(\text{MLP}(\tilde{u}_{v}^{l})+\tilde{u}_{v}^{l}),
\end{align}
where $u_v^0=f_v$, $u_a^0=f_a$, $u_v^{l-1} \in \mathbb{R}^{S\times T\times C}$ are the output visual embeddings of the previous LSCM block, $\text{LN}(\cdot)$ denotes Layer Normalization~\cite{ba2016layer}, $\text{MHA}(q,k,v)$ is multi-head attention~\cite{vaswani2017attention} with query $q$, key $k$, value $v$, and MLP is a multi-layer perceptron. Audio Self-Attention is applied in the same way to $u_a^{l-1}$ to obtain the audio output $\tilde{u}_{a}^{l}$.

\xhdr{Audio-Visual Cross-Attention.}
The visual and audio streams are so far processed separately. To enhance visual features with audio and vice versa, we use an Audio-Visual dual Cross-Attention,
\begin{align}
\label{eqn:2}
    \widehat{u}_{v}^{l} &= \text{LN}(\text{MHA}(\tilde{u}_{v}^{l},\tilde{u}_{a}^{l},\tilde{u}_{a}^{l})+\tilde{u}_{v}^{l}),\\
    \widehat{u}_{v}^{l} &= \text{LN}(\text{MLP}(\widehat{u}_{v}^{l})+\widehat{u}_{v}^{l}),
\end{align}
where $\widehat{u}_{v}^{l}$ are audio-enhanced visual embeddings. Visual-enhanced audio embeddings $\widehat{u}_{a}^{l}$ are obtained in the same way via Audio-Visual Cross-Attention given $\tilde{u}_{a}^{l}$ and $\tilde{u}_{v}^{l}$.

\subsubsection{Short-term Inter-speaker Modeling (SIM)}
For a given moment in the video, the speaking activity of a target person is more coordinated with other speakers in closer frames~\cite{richardson2008synchrony}, so 
the model should capture local temporal inter-speaker relationships. 
To do this, we employ a small \textbf{Inter-speaker Convolutional Network},
\begin{align} \label{eq3}
u_{v}^l &= \text{MLP}(\text{LN}(\text{Conv}_{s\times k}(\widehat{u}_{v}^{l})))+\widehat{u}_{v}^{l},\\
u_{a}^l &= \text{MLP}(\text{LN}(\text{Conv}_{s\times k}(\widehat{u}_{a}^{l})))+\widehat{u}_{a}^{l},
\end{align}
where visual embeddings $u_{v}^l\in \mathbb{R}^{S\times T\times C}$ and audio embeddings $u_{a}^l\in \mathbb{R}^{S\times T\times C}$  are the output of the $l$-th LSCM block that will be passed to the next block. $k$ is the temporal length of the receptive field and $s$ is the number of speakers considered. Explicitly modeling inter-speaker context in nearby frames enables cross-frame inter-speaker information exchange. Our SIM module with a short temporal receptive field can help capture local dynamic patterns in interactions.

\subsection{Training and Inference}
Following~\cite{carion2020end,al2019character,szegedy2015going}, we train our model with multiple supervisions utilizing $u^N \in \mathbb{R}^{S\times T\times 2C}$ (Sec.~\ref{sec:lscm}) and $u^i$ derived from each intermediate block of LSCM. 
For each $u^i$, a fully-connected (FC) layer is applied, yielding prediction results $\hat{R}^i \in \mathbb{R}^{T}$ corresponding to the target speaker for each frame. All FC layers share their parameters. The overall loss function is $L = \sum_{i=1} ^{N} \text{CrossEntropy}(\hat{R}^i, R)$ where $R$ is the ground-truth. 
During inference, we can reduce the computation by a factor of $S$ by reusing speakers' features extracted by the visual encoder, which uses the most FLOPs.

\subsection{Implementation details} 

Following~\cite{tao2021someone}, our visual encoder consists of a 3D convolutional layer,  ResNet-18~\cite{he2016deep}, and a visual temporal convolution network (V-TCN)~\cite{lea2016temporal}. 

Our audio encoder is the proposed VGGFrame initialized with VGGish~\cite{chen2020vggsound} weights pretrained on AudioSet~\cite{gemmeke2017audio}.

We sample $S=3$ speakers and $T=200$ frames. In SIM, $s$ is set to 3, which is the same as $S$, and $k$ is set to 7 frames. The face crops are resized to $112\times112$. Visual augmentation includes randomly resized cropping, horizontal flipping, and rotation. For audio augmentation, another audio signal is randomly chosen from the rest of the training set and added as noise to the target audio.
We train LoCoNet with Adam~\cite{kingma2014adam} for 25 epochs on 4 RTX6000 GPUs with batch size  4 using PyTorch~\cite{paszke2019pytorch}. The learning rate is $ 5 \times 10^{-5}$ and reduced by 5\% each epoch.
\section{Experimental Setup}
\label{sec:experiment}


\subsection{Datasets}

\xhdr{AVA-ActiveSpeaker}~\cite{roth2020ava} is a standard benchmark for ASD, consisting of 262 videos from Hollywood movies with 3.65 million frames and 5.3 million face crops. 
Following~\cite{liao2023light,min2022learning,min2022intel,datta2022asd,alcazar2022end,jiang2023target}, we evaluate on the validation set.

\xhdr{Talkies~\cite{leon2021maas}} is an in-the-wild ASD dataset with 23,507 face tracks extracted from 421,997 labeled frames. It focuses on challenging cases with more speakers, diverse actors and scenes, and more off-screen speech.

\xhdr{Ego4D~\cite{grauman2022Ego4D}}'s Audio-Visual benchmark has 3,670 hours of egocentric video of unscripted daily activities in many environments. It includes many challenges that complement  exocentric benchmarks, including unusual viewpoints and speakers who are  off-screen.

\subsection{Evaluation Metric}

Following~\cite{tao2021someone,alcazar2020active,kopuklu2021design,leon2021maas,alcazar2022end}, we use the official ActivityNet~\cite{caba2015activitynet} evaluation tool to compute mean average precision (mAP) and evaluate on the AVA-ActiveSpeaker~\cite{roth2020ava}. We also compute AUC~\cite{bradley1997use} as another evaluation metric using Sklearn~\cite{scikit-learn}. We use mAP to evaluate on Talkies~\cite{leon2021maas} and Ego4D~\cite{grauman2022Ego4D}.

\section{Results and Analysis}

We first compare the proposed method ~\ModelName~with previous state-of-the-art methods on multiple datasets and challenging scenarios. Then we validate our hypotheses of long-term intra-speaker and short-term inter-speaker modeling. Finally, following \cite{kopuklu2021design, alcazar2020active,tao2021someone,liao2023light,zhang2021unicon}, we conduct extensive ablations on each component of \ModelName~(on AVA-ActiveSpeaker~\cite{roth2020ava} unless otherwise noted).

\subsection{Comparison with State-of-the-Art}

In this section, we compare our approach with state-of-the-art methods on the three datasets.

\xhdr{AVA-ActiveSpeaker.} From Table~\ref{table:sota_map}, end-to-end methods exhibit fewer FLOPs while maintaining competitive mAP compared to multi-stage methods. The higher FLOPs of multi-stage methods stem from their sequential-inference processing strategy (Sec. \ref{sec:relatedwork}) of stacking multiple neighboring frames centered at time $t$.
\ModelName~achieves a 95.2\%  mAP, surpassing the best-performing end-to-end ASD method Light-ASD~\cite{liao2023light} by 1.1\% while using modest average-FLOPs.
Moreover, \ModelName~outperforms previous state-of-the-art multi-stage method SPELL+~\cite{min2022learning} by 0.3\% with about 32\% fewer parameters and over 38$\times$ fewer average-FLOPs.

\begin{table}[tb]
\small
\setlength{\tabcolsep}{2pt}
\centering
\begin{tabular}{lccccc}

\toprule
\multirow{2}{*}{\textbf{Method}} & \multirow{2}{*}{\textbf{Vid. Enc.}} & \multirow{2}{*}{\textbf{Params}} & \textbf{\textit{average}-} & \multirow{2}{*}{\bf mAP} &  \multirow{2}{*}{\bf AUC} \\
&&& \bf FLOPs &&\\
\midrule
\multicolumn{6}{l}{\textit{Multi-Stage}} \\
ASC ~\cite{alcazar2020active} & R-18~\cite{he2016deep} & $23.0M$ & $1.0G$ & 87.1 & - \\
MAAS ~\cite{leon2021maas}  & R-18 & $23.0M$ & $1.6G$ &  88.8 & - \\
ASDNet ~\cite{kopuklu2021design}  & 3DRNext-101  & $51.0M^\dag$ & $13.2G^\dag$ & 93.5 & -\\
SPELL ~\cite{min2022learning} & R-18+TSM  & $23.5M^{\dag}$ & $8.7G^{\dag}$ & 94.2 & -\\
SPELL+ ~\cite{min2022learning} & R-50+TSM  & $51.23M^{\dag}$ & $19.6G^{\dag}$ & 94.9 & -\\
\midrule
\multicolumn{6}{l}{\textit{End-to-End}} \\
UniCon ~\cite{zhang2021unicon} & R-18  & $23.8M$ & $3.0G^{\dag}$ &  92.2 & 97.0 \\
TalkNet ~\cite{tao2021someone}   & R-18+VTCN & $15.7M$ & $0.51G$ &  92.3 & 96.8  \\
ASD-Trans ~\cite{datta2022asd} & R-18+VTCN  & $15.0M$ & $0.55G$ & 93.0 & - \\ 
EASEE ~\cite{alcazar2022end} & 3D R-50  & $26.8M^{\dag}$ & $4.3G^{\dag}$ & 94.1 & - \\
Light-ASD \cite{liao2023light} & (2+1)D Conv & \bm{$1.02M$} & \bm{$0.2G$} & 94.1 & - \\
TS-TalkNet \cite{jiang2023target} & R-18+VTCN & $36.8M$ & $2.3G$ & 93.9 & - \\ 
\ModelName (Ours) & R-18+VTCN  & $34.3M$ & $0.51G$ & \textbf{95.2} & \textbf{98.0}\\
\hline
\end{tabular}
\caption{\textbf{Comparison with SOTAs on AVA-ActiveSpeaker}. \text{3DRNext} denotes 3D ResNext~\cite{xie2017aggregated}. \text{R} denotes 2D ResNet~\cite{he2016deep}. \textit{average}-FLOPs represents the averaged FLOPs needed to process a single face crop. $^\dag$ denotes our estimates based on 
their visual encoders. Most methods incur higher costs by extracting features for each frame through stacking multiple adjacent frames (\ie, 11 in SPELL). \ModelName~achieves the highest mAP with modest FLOPs.}
\label{table:sota_map}
\end{table}

\begin{table}[t]
\centering
\begin{tabular}{cccc}
\toprule
\multirow{2}{*}{\bf Method} & \multicolumn{2}{c}{\bf Train Set} & \multirow{2}{*}{\bf mAP} \\
\cmidrule(lr){2-3}
& AVA & Talkies & (\%)\\
\midrule
MAAS~\cite{leon2021maas} &	\cmark&	\xmark&	79.7\\
EASEE~\cite{alcazar2022end} & \cmark & \xmark&	86.7\\
\textbf{LoCoNet} &	\cmark&	\xmark&	\textbf{88.4}\\
\midrule
EASEE~\cite{alcazar2022end} &	\xmark&	\cmark&	93.6\\
light-ASD\cite{liao2023light} & \xmark & \cmark & 93.9 \\
\textbf{LoCoNet}	&\xmark&	\cmark&	\textbf{96.1}\\
\midrule
EASEE~\cite{alcazar2022end}  &\cmark&	\cmark&	94.5\\
\textbf{LoCoNet} & \cmark &	\cmark&	\textbf{97.2}\\
\bottomrule
\end{tabular}
\caption{\textbf{Comparison on Talkies dataset} under three training settings: train on AVA-ActiveSpeaker alone, train on Talkies  alone, or train on AVA-ActiveSpeaker and finetune on Talkies. 
}
\label{table:talkies}
\end{table}

\xhdr{Talkies set.}
We evaluate \ModelName~with other methods under three training settings: (i) AVA-ActiveSpeaker, (ii) Talkies, and (iii) pretrained on AVA-ActiveSpeaker and finetuned on Talkies.
As shown in Table~\ref{table:talkies}, \ModelName~outperforms EASEE by 1.7\%, 2.5\%, and 2.7\% in these three settings, respectively. It also outperforms light-ASD by 2.2\% when both models are trained on Talkies only.

\noindent \textbf{Ego4D dataset.}
We evaluate our method on Ego4D Audio-Visual benchmark~\cite{grauman2022Ego4D}. 
\ModelName~achieves 68.4\% mAP, outperforming TalkNet and Challenge Winner~\cite{min2022intel} (SPELL~\cite{min2022learning}-based) by 16.7\% and 7.7\% respectively. Egocentric videos, characterized by constant camera motion, lower clarity, and more complex scenes 
compared to exocentric videos, demonstrate the potential of the proposed Long-Short Context Modeling in real-life scenarios. Our approach  benefits from capturing both multi-speaker interaction and single-speaker behavior.

\begin{table}[bt]
\centering
\begin{tabular}{ccc}
\toprule
\bf Method &  \bf mAP (\%) \\
\midrule
TalkNet~\cite{tao2021someone}  & 51.7\\
Challenge Winner~\cite{min2022intel} & 60.7 \\
\textbf{\ModelName} & \textbf{68.4}\\
\bottomrule
\end{tabular}
\caption{\textbf{Comparison on Ego4D dataset.}  The challenge winner is not specifically optimized for ASD  but the large improvement ($+7.7\%$) still shows the strong generalizability of \ModelName. The result of TalkNet was obtained by us with their released code.
}
\label{table:ego}
\end{table}

\subsection{Challenging Scenario Evaluation}

\noindent \textbf{Quantitative analysis.}

We report the performance of \ModelName~on AVA-ActiveSpeaker under different face sizes:
(i) Small: faces with width less than 64 pixels; (ii) Medium: faces with width between 64 and 128 pixels;  (iii) Large: faces with width larger than 128 pixels. We also study the effect of the number of visible faces in a video frame (1, 2, or 3).

The  results, along with the portions of each category, are shown in Tables~\ref{table:challenging} and ~\ref{table:challenging_num} respectively. \ModelName~consistently performs the best across all scenarios, exhibiting the most significant improvement in the challenging multi-speaker case: +3.0\% for 3 faces. 
This suggests that our method more

effectively models both the target speaker's speaking pattern and the interactions of context speakers, allowing accurate inference of the speaking activity of the target person.

\begin{table}
\centering
\begin{tabular}{cccc}
\toprule
\multirow{3}{*}{Method}& \multicolumn{3}{c}{Face Size}\\
\cmidrule(lr){2-4}
& \makecell{Small \\ (18\%)} & \makecell{Medium \\ (30\%)} & \makecell{Large \\ (52\%)} \\
\midrule
ASC~\cite{alcazar2020active}  & 56.2 & 79.0 & 92.2 \\
MAAS~\cite{leon2021maas}  & 55.2 & 79.4 & 93.0 \\
TalkNet~\cite{tao2021someone}  & 63.7 & 85.9 & 95.3 \\
ASDNet~\cite{kopuklu2021design}  & 74.3 & 89.8 & 96.3 \\
EASEE~\cite{alcazar2022end} & 75.9 & 90.6 & 96.7 \\
light-ASD~\cite{liao2023light} & 77.5 & 91.2 & 96.5 \\

\textbf{\ModelName} & \textbf{77.8} & \textbf{93.0} & \textbf{97.3}\\
\bottomrule
\end{tabular}

\caption{\textbf{Results as a function of   face size.} \ModelName~achieves the highest mAP among all face sizes. 
}
\label{table:challenging}
\end{table}

\begin{table}
\centering
\begin{tabular}{cccc}
\toprule
\multirow{2}{*}{Method} & \multicolumn{3}{c}{\# Faces}\\
\cmidrule(lr){2-4}
& \shortstack{1 \\ (45\%)} & \shortstack{2 \\ (33\%)} & \shortstack{3 \\ (11\%)} \\
\midrule
ASC~\cite{alcazar2020active} & 91.8 & 83.8 & 67.6 \\
MAAS~\cite{leon2021maas} & 93.3 & 85.8 & 68.2 \\
TalkNet~\cite{tao2021someone} & 95.4 & 89.6 & 80.3\\
ASDNet~\cite{kopuklu2021design} & 95.7 & 92.4 & 83.7 \\
EASEE~\cite{alcazar2022end} & 96.5 & 92.4 & 83.9 \\
light-ASD~\cite{liao2023light} & 96.2 & 92.6 & 84.4 \\
\textbf{\ModelName} & \textbf{97.0} & \textbf{94.6} & \textbf{87.4}\\

\bottomrule
\end{tabular}
\caption{\textbf{Results as a function of visible faces in the  scene.} Larger improvements are observed on more challenging cases (\ie, 3 faces).}
\label{table:challenging_num}
\end{table}

\begin{figure*}[tb]
\centering
\includegraphics[width=\textwidth]{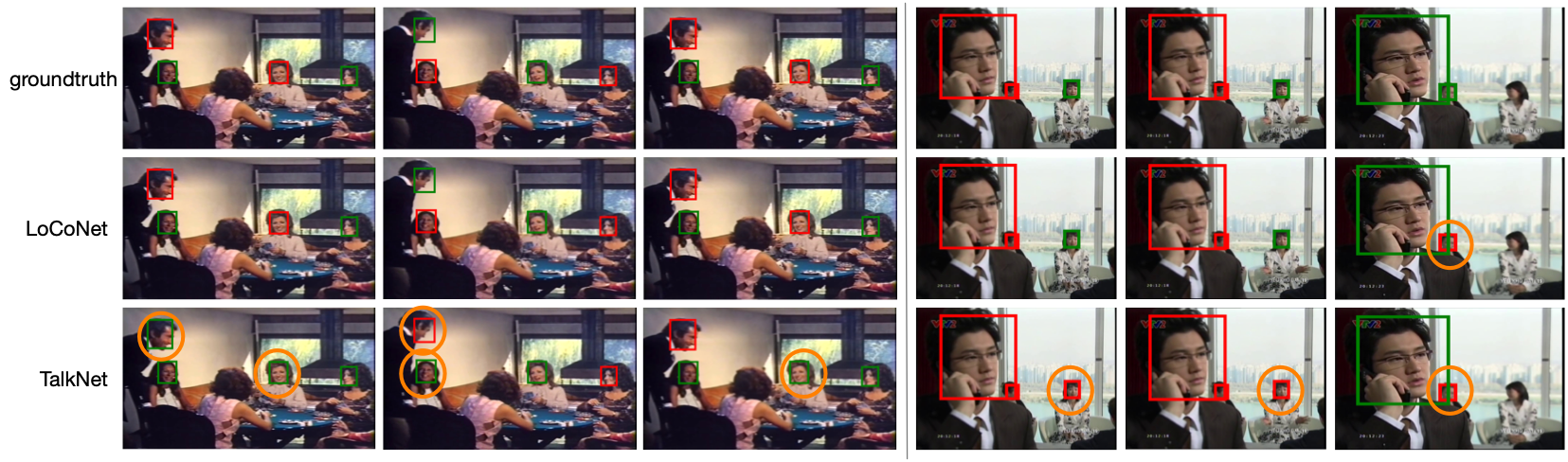}
\caption{\textbf{Results comparison of \ModelName~and TalkNet on challenging scenarios of AVA-ActiveSpeaker.} \textcolor{red}{Red} boxes denote not-active speaker. \textcolor{ao}{Green} box denote active speaker. \textcolor{orange}{Orange} circles refer to false predictions. The video on the left shows a multi-people conversation with four speakers, and separate conversation of two. The video on the right shows an active speaker with a small face. Both scenes are challenging, and \ModelName~predicts accurately in most cases. }
\label{fig:quality}
\end{figure*}

\noindent \textbf{Qualitative analysis.}
Fig~\ref{fig:quality} visualizes the results of \ModelName~and TalkNet~\cite{tao2021someone} on AVA-ActiveSpeaker~\cite{roth2020ava} with the groundtruth labels. The video on the left shows four visible speakers talking, posing a challenge in distinguishing the active speaker amid multiple discussions in the same scene. \ModelName~accurately locates the active speakers in this case, whereas TalkNet fails to recognize some of them. The first two columns of the video on the right shows a woman with a very small visible face as the active speaker, while the man with a large visible face is not speaking. TalkNet fails to locate the active speaker while \ModelName~succeeds.  By combining long-term intra-speaker context to compare the speaking pattern of each individual and short-term inter-speaker context to examine the conversations, our approach better overcomes this challenging speaking scenarios. However, in the last column, both methods fail to recognize the active speaker at the back. This scenario is especially challenging as the two active speakers are in separate conversations with one being far less salient than the other, making it difficult to infer the speaking activity of the less salient speaker.

\subsection{Attention visualizations of LIM and SIM}
\begin{figure}
\centering
\includegraphics[width=\linewidth]{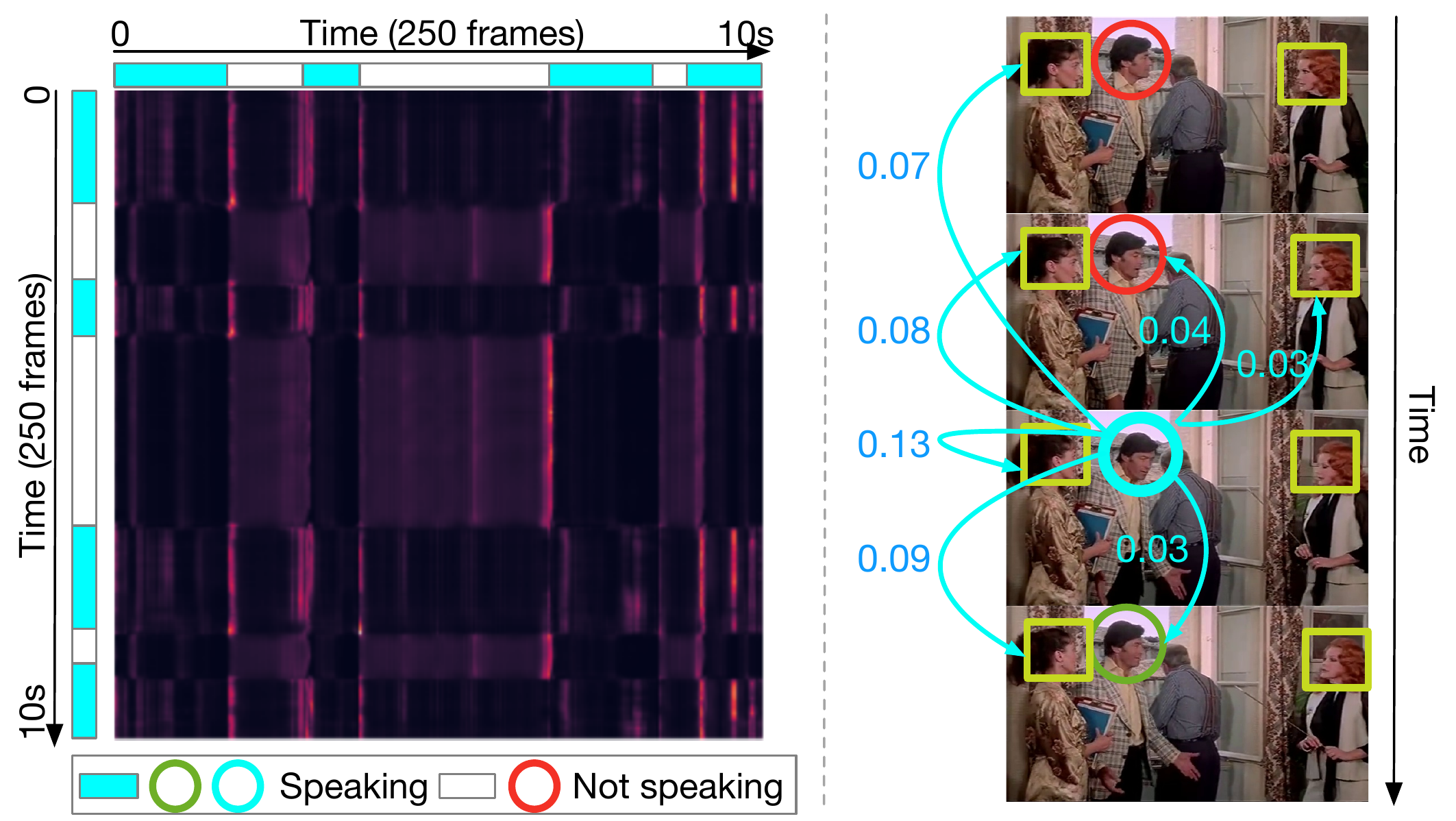}
\caption{\textbf{Visualization of LIM and SIM.} Left: attention weights of one speaker in LIM, showing the ability to capture long-term intra-speaker context.  Right: to predict the speaking activity of the {\color{cyan} target speaker at target frame}, we show the information drawn from {\color{red} not-speaking target} and {\color{asparagus} speaking target} and {\color{arylideyellow} context speakers} at nearby frames. }
\label{fig:lim-sim}
\end{figure}

We  next visualize    the effectiveness of Long-term Intra-speaker Modeling (LIM) and Short-term Inter-speaker Modeling (SIM). The left part of Fig.~\ref{fig:lim-sim} visualizes the attention weights across different frames of a single speaker in LIM. It is evident that speaking and non-speaking activities are distinctly separated, with clear boundaries when speaking activities change. This verifies that LIM contributes to  accurate speaking activity detection. The right part of Fig.~\ref{fig:lim-sim} shows the portions of information (measured by L2-norm, as convolutions are used for SIM) drawn by the target speaker at target frame from all speakers at nearby frames (and information flows with very small portions are not shown). The woman on the left gradually turns her face to the target speaker,  which is the most indicative sign that the target speaker has started speaking. The distribution of information flow reveals that SIM  infers from the behaviors of context speakers, assigning more attention to the woman on the left. 

\subsection{Context Modeling Analysis}

\xhdr{Does Intra-speaker Modeling require long-term?} We train \ModelName~by keeping the number of  speakers $S$ as 1, and varying the temporal length of input frames $T$ from 20 to 400. Table~\ref{tab:sub_first} indicates that the network performs worst when trained with the shortest video segments of 20 frames (0.8 sec). Performance  improves as video segments become longer, with a 5.0\% mAP increase at 100 frames (4 sec) and an additional 0.4\% at 200 frames. This underscores the importance of long-term temporal context in intra-speaker modeling, aligning with findings from TalkNet~\cite{tao2021someone} and ASC~\cite{alcazar2020active} that long-term temporal context provides better evidence of a speaking episode. 
 
We found 200 frames to be a good balance between performance and memory cost.

\begin{table}
\centering
\begin{tabular}{cccccc}
      \toprule
      \# Frames & 20 & 100 & 200 & 300 & 400\\
      \midrule
      mAP (\%) & 87.8 & 92.8 & \textbf{93.2} & 93.2 & OOM\\
      \bottomrule
   \end{tabular}
   \caption{\textbf{Temporal Length} in Long-term Intra-speaker Modeling. We set $S=1$ so the context is intra-speaker only.}
   \label{tab:sub_first}
\end{table}

\begin{table}
\centering
\begin{tabular}{cccc}
     \toprule
      Speakers & 1 & 2 & 3 \\
      \midrule
      mAP (\%) & 93.2 & 94.6 & \textbf{95.2} \\
      \bottomrule
   \end{tabular}
   \caption{\textbf{Number of Speakers $S$} in Short-term Inter-speaker Modeling. We set the number of frames $T$ to 200.}
   \label{tab:sub_second}
\end{table}

\begin{table}
\centering
\begin{tabular}{cccccc}
      \toprule
      Receptive Field & 1 & 7 & 15 & 31 & 101 \\
      \midrule
      GFLOPs & \textbf{0.28} & 0.98 & 1.93 & 3.82 & 12.1 \\
      \midrule
      mAP(\%) & 94.5  & \textbf{95.2} & \textbf{95.2} & 95.0 & 94.9 \\
      \bottomrule
   \end{tabular}
   \caption{\textbf{Temporal Receptive Field $k$} in SIM, indicating  how many neighboring frames each speaker can explicitly query on in the SIM module. $S$ is set as 3 and $T$ as 200.}
   \label{tab:receptive}
\end{table}

\xhdr{Is Short-term Inter-speaker Modeling Sufficient?}
We first validate the importance of inter-speaker context. Keeping the temporal length at 200 frames, we vary the number of speakers $S$ from 1 to 3. Larger $S$ values were not considered because $>$ 99\% of videos in AVA-ActiveSpeaker~\cite{roth2020ava} have at most 3 speakers in the same scene. Table~\ref{tab:sub_second} demonstrates that performance increases with more speakers included in training. This supports our hypothesis that modeling multiple speakers in the inter-speaker context is necessary for ASD. 

Next, with $S=3$, we vary the temporal receptive field $k$ of SIM from 1 (40 msec) to 101 (4 sec). Table~\ref{tab:receptive} shows a 0.7\% performance increase when increasing $k$ from 1 to 7, confirming our assumption that motion in short-term inter-speaker context is more valuable than in-frame context. Performance saturates with further increases in the receptive field, and the computation cost increases drastically. This reinforces our hypothesis in Sec.\ref{sec:intro} that short-term inter-speaker modeling is sufficient.

\subsection{Other Ablations}

\xhdr{Does each component of \ModelName~help?}
In Table~\ref{table:component}, replacing ResNet-34~\cite{he2016deep} with VGGFrame as the audio encoder enhances mAP by 1.6\%, showing the effectiveness of  an audio encoder pretrained on audio datasets compared to a common vision encoder. With VGGFrame as audio encoder, adding 3 Long-term Intra-speaker Modeling (LIM) modules increases the performance by 1.3\%, while 3 Short-term Inter-speaker Modeling (SIM) modules adds 1.2\%.  Both together improve mAP by 2.4\%.

\begin{table}[]
\centering
\begin{tabular}{ccccc}
\toprule
\multicolumn{4}{c}{\textit{Ablation Settings}} & mAP \\
R-34 & VGGFrame & LIM & SIM & (\%) \\
\midrule
 \cmark& & & & 91.2 \\
&\cmark & & & 92.8 \\
&\cmark & \cmark & & 94.1\\
&\cmark & & \cmark & 94.0 \\
&\cmark & \cmark & \cmark & \textbf{95.2} \\
\bottomrule
\end{tabular}
\caption{\textbf{Ablations on components of \ModelName.} 
We study the efficacy of audio encoder (ResNet-34 or VGGFrame), Long-term Intra-speaker Modeling (LIM) and Short-term Inter-speaker Modeling (SIM). LIM and SIM are stacked 3 times if applied.}
\label{table:component}
\end{table}

\xhdr{Convolution versus Window Self-Attention in SIM.}
\begin{table}[]
\centering
\begin{tabular}{lccc}
\toprule
Design  & mAP(\%) \\
\midrule
Convolution & \textbf{95.2} \\
Window Self-Attention & 94.4\\
\bottomrule
\end{tabular}
\caption{\textbf{Ablation on designs of Short-term Inter-speaker Modeling}. We implement SIM using Convolution or Window Self-Attention (Eqn.~\ref{eq3}). We keep the reception field the same (\ie, 7 frames) and stack it  three times ($N=3$).}
\label{table:sim}
\end{table}
Besides Convolution, we also try Window Self-Attention to capture local patterns for SIM (Eqn.~\ref{eq3}). In Table~\ref{table:sim}, Convolution outperforms Window Self-Attention by 0.8\% mAP, highlighting the superiority of Convolution in modeling local patterns of speakers' interaction.

\begin{table}[]
\centering
\begin{tabular}{lcccccc}
\toprule
    $N$ & 0 & 1 & 2 & 3 & 4 & 5\\
      \midrule
      mAP(\%) & 92.8 & 94.3 & 95.0 & \textbf{95.2} & 95.0 & 95.0\\
\bottomrule
\end{tabular}
\caption{\textbf{Ablation on the number of blocks in LSCM ($N$).} $N=0$ refers to \ModelName~with no context modeling.}
\label{table:lscm}
\end{table}
\xhdr{Number of blocks in LSCM.} We vary the number of blocks $N$ in Long-Short Context Modeling (Sec.~\ref{sec:lscm}). Adding 1 block of LSCM yields a performance gain of 1.5\%, showing the effectiveness of LIM and SIM. Adding two more blocks further  increases  by 0.7\%, but results saturate at $N=3$.

\section{Conclusion}
\label{sec:conclusion}

In this work, we observe that  speaker activity can be more efficiently inferred from long-term intra-speaker context and short-term inter-speaker context.
We thus design an end-to-end long-short context ASD framework that uses self-attention and cross-attention mechanisms to model long-term intra-speaker context and a convolutional network to model short-term inter-speaker context.
With a simple backbone network, our method achieves state-of-the-art performance on 3 mainstream ASD benchmarks and significantly outperforms previous state of the art methods  by 7.7\% on Ego4D.
We also show that in challenging scenarios where multiple speakers are in the same scene or speakers have small faces, our proposed method also outperforms previous methods. All of these results show the robustness and effectiveness of our method.
Similar to existing long-term ASD methods, our method utilizes  8 sec of context. Future work should study how to implement larger contexts. Additionally, enhancing ASD in egocentric datasets could include adding other modalities,
such as eye gaze.

\section*{Acknowledgements}
The authors gratefully acknowledge
Prof.~David Crandall's guidance
and feedback on earlier versions of this work. 
This work was supported in part by the National Science Foundation under award DRL-2112635 to the  AI Institute for Engaged Learning, Sony Faculty Innovation Award, Laboratory for Analytic Sciences via NC State University, and ONR Award N00014-23-1-2356. Any opinions, findings, and conclusions or recommendations expressed in this material are those of the author(s) and do not necessarily reflect the views of the National Science Foundation.

{
    \small
    \bibliographystyle{ieeenat_fullname}
    \bibliography{main}
}


\end{document}